# CA-CentripetalNet: A novel anchor-free deep learning framework for hardhat wearing detection

Zhijian Liu[a], Nian Cai*,[a,b], Wensheng Ouyang[a], Chengbin Zhang[a], Nili Tian[a], Han Wang[b, c]

[a] *School of Information Engineering, Guangdong University of Technology, Guangzhou, 510006, China*

[b] *State Key Laboratory of Precision Electronic Manufacturing Technology and Equipment, Guangdong University of Technology, Guangzhou 510006, P.R. China*

[c] *School of Mechanical and Electrical Engineering, Guangdong University of Technology, Guangzhou 510006, P.R. China*

*corresponding author: cainian@gdut.edu.cn

**Abstract**

Automatic hardhat wearing detection can strengthen the safety management in construction sites, which is still challenging due to complicated video surveillance scenes. To deal with the poor generalization of previous deep learning based methods, a novel anchor-free deep learning framework called CA-CentripetalNet is proposed for hardhat wearing detection. Two novel schemes are proposed to improve the feature extraction and utilization ability of CA-CentripetalNet, which are vertical-horizontal corner pooling and bounding constrained center attention. The former is designed to realize the comprehensive utilization of marginal features and internal features. The latter is designed to enforce the backbone to pay attention to internal features, which is only used during the training rather than during the detection. Experimental results indicate that the CA-CentripetalNet achieves better performance with the 86.63% mAP (mean Average Precision) with less memory consumption at a reasonable speed than the existing deep learning based methods, especially in case of small-scale hardhats and non-worn-hardhats.

## 1. Introduction

Nowadays, a large number of employees work in the construction sites [1], who are in the danger at all times [2, 3]. Statistically, construction industry accident is one of the main cause of traumatic brain injury, which is likely to lead to long-term disability or even death [4, 5]. The safety hardhat, a personal protective equipment, can effectively reduce fatal brain injuries caused by safety accidents in the construction sites, which can protect the workers and even save their lives. Thus, workers are required to wear safety hardhats according to safety regulations in various countries [6, 7]. However, some workers may be careless with safety regulations. In the past, implementation of safety regulations were reliant on inspectors' manual monitoring and reporting, which brought high labor cost and was difficult to ensure the real-time monitoring. Therefore, it is of significance for promoting the research on the automatic detection of hardhat wearing to strengthen the safety management of the construction site.

Early related studies mainly include sensor based methods [8,9,10] and traditional computer vision based methods [11,12,13], which meet with many problem in practical applications. Recently, with the rapid development of deep learning, a series of studies have introduced convolutional neural networks into hardhat wearing detection and achieved some breakthroughs [14-22], which can be roughly divided into two categories, i.e. multi-step and single-step methods. Multi-step methods [14-16] implement the hardhat wearing detection by means of human or face detection and subsequent hardhat wearing detection, which assume that a correctly worn hardhat is always on the top of a human body or face. Nath et al. [14] directly employed the YOLOv3 to detect and crop the workers from the image, and then employed a classification network to recognize whether the worker was wearing a hardhat. Chen et al. [15] simultaneously used an OpenPose for detecting keypoints of the human body and a YOLOv3 for detecting the hardhat. Then, the geometric relationship between the hardhat region and the keypoint of the neck was employed to determine whether the hardhat was worn correctly. A face-to-helmet regression model [16] is proposed to first locate the head of the worker based on small face recognition, and then to use a DenseNet classifier to identify whether the worker is wearing a hardhat. However, the pedestrian detection-based methods [14,15] suffer from occlusion issues [16] and the face detection-based method [16] can not detect workers with their back to the camera. In these multi-step methods, the latter detection for the hardhat does not function, if the detector for the first step fails to detect the human body or the face due to the occlusion or the camera angle. For the safety monitoring, this missed detection is usually more fatal than a false detection in practice. Therefore, the multi-step methods are unreliable in the complex environment of the construction sites.

To solve the problem existed in the multi-step methods, some researchers [17-22] have provided single-step methods to simultaneously identify whether there is a worker and he is wearing a hardhat in a framework. Fang et al. [17] proposed a Faster R-CNN[18] based method to detect the entire construction workers and to classify whether they were wearing hardhats. Wu et al. [19] proposed a SSD [20] framework with reverse progressive attention to identify whether the worker is wearing a hardhat with which color. Wang et al. [21] designed a light detector for hardhat wearing detection, which incorporated a top-down module and a residual-block-based prediction module into a MobileNet [22]. Huang at el. [23] introduced several ideas into the YOLOv3 [24] for hardhat wearing detection, which involves multi-scale fusion, multi-scale prediction and CIou loss function. Then, the hardhat color was distinguished by the pixel feature statistics of the detected hardhat. Wang et al. [25] proposed cross stage YOLOv3 to real-time detect the worn hardhat, which introduced the cross stage partial network with a spatial pyramid pooling (SPP) module to reinforce the feature extraction network of YOLOv3. Wang et al. [26] evaluated eight existed YOLO networks for detecting whether the hardhat or vest was worn and which color was the worn hardhat. They found out that the YOLOv5x had achieved the best mAP (mean Average Precision) for hardhat wearing detection. These existing single-step methods are anchor-based detection networks, which consume much memory. Also, they have high sensitivity to the hyperparameters of the networks, which results in poor generalization [27-28]. These limitations will influence their practical applications, due to the complex environment of the construction sites and the limited memory resource of the video surveillance system.

Compared with the anchor-based networks, the anchor-free network has the advantages of high efficiency, simple structure and rich generalization, which has attracted more and more researchers in the field of computer vision [27-32]. Corner prediction based detection network [33-35], an important anchor-free networks, predicts two groups of heatmaps representing the positions and confidence scores of top-left corner and the bottom-right corner of the objects, respectively. This means that it is sensitive to marginal information[34]. In construction sites, this sensitivity will make it mis-distinguish many objects with similar marginal features to the hardhats as the hardhats, which reduces its generalization.

To further improve the detection accuracy for hardhat wearing detection with a reasonable memory resource, we design an anchor-free deep learning framework for hardhat wearing detection, which incorporates a training module into the framework to deal with the above sensitivity problem of the previous anchor-free networks. For simplicity, the novel framework is termed as CA-CentripetalNet, which determines whether the workers in the construction site are wearing hardhats and distinguish which colors are the worn hardhats. The vertical-horizontal corner pooling is proposed to achieve more significant corner information compared with the traditional corner pooling, which combines marginal features with internal features. An extra prediction head is designed to excellent extract internal features, which is called bounding constrained center attention module. It is noted that this module is used only for training to enforce the backbone to pay more attention to the internal information.

To summarize, our work provides three contributions.

(1) Previous single-step deep learning methods for hardhat wearing detection are anchor-based, which have the disadvantages of high memory consumption and poor generalization. In this paper, a novel anchor-free framework is proposed for hardhat wearing detection, which is a single-step and end-to-end deep learning method. Compared with the previous hardhat wearing detection networks, the proposed framework has the advantages of better detection accuracy and less memory consumption.

(2) If previous anchor-free networks are directly employed for hardhat wearing detection, they have a high risk of mis-distinguishing many objects as the hardhats, which have similar marginal features to the hardhats. Here, we design a vertical-horizontal corner pooling scheme to comprehensively utilize marginal features and internal features of the objects, which is beneficial to better characterize the hardhats with more corner information compared with previous anchor-free networks.

(3) To incorporate the internal features into the proposed framework without the increase of detection computational burden, a training module is designed to predict an extra center point and a bounding constrained vector of the object, which is not involved for detection. It is called bounding constrained center attention and only works in training. It forces the framework to pay attention to the interior of the object, which can improve the ability to extract the internal features.

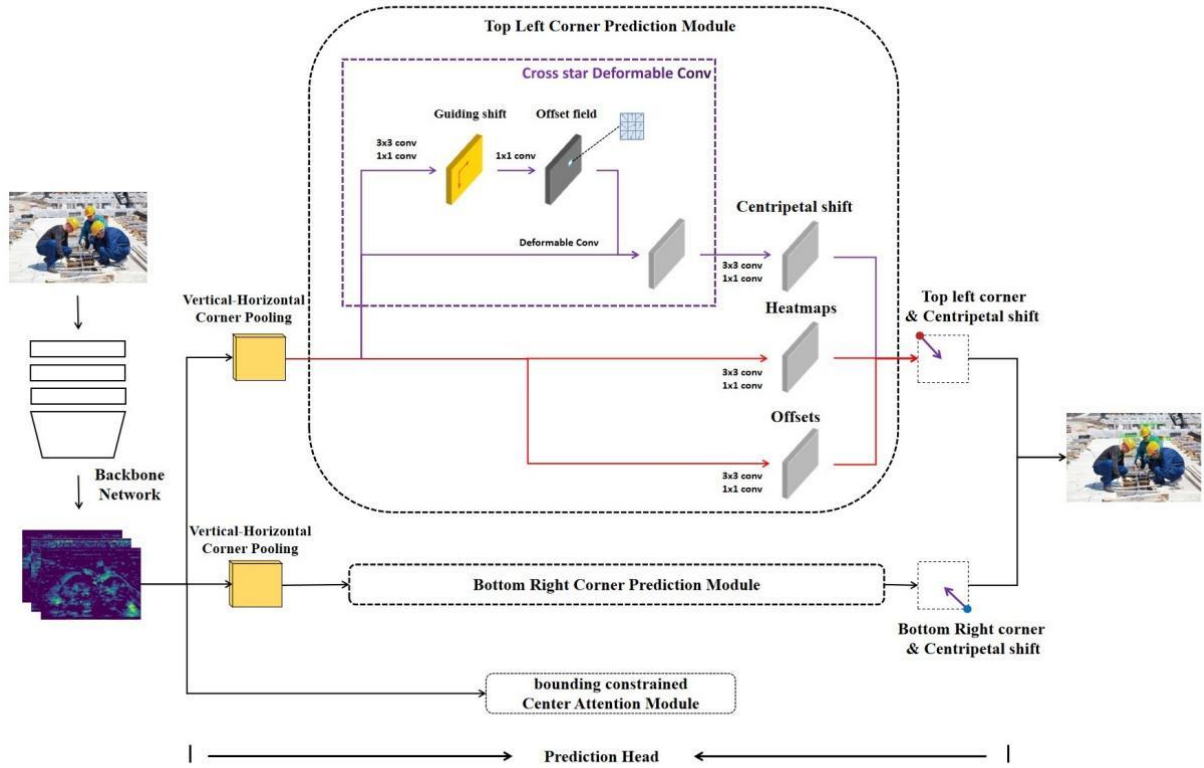

Fig. 1. CA-CentripetalNet for hardhat wearing detection.

## 2. Methodologies

### 2.1. Architecture of the proposed framework

As illustrated in Fig. 1, the proposed framework called CA-CentripetalNet is composed of a backbone network and a prediction head. The backbone network is to extract and aggregate image features from the surveillance image of the construction site. The prediction head consists of three network branches, two of which have the similar structures including a vertical-horizontal corner pooling and a corner prediction module for top-left corners and bottom-right corners, respectively. And a bounding constrained center attention module is designed as another network branch. It is noted that this network branch works during the framework training rather than during the detection.

The vertical-horizontal corner pooling is designed to focus the local features located in the interior and horizontal margin of the object on the corner position, which can improve the accuracy of corner prediction. The corner prediction module predicts a heatmap representing the locations of corners of different categories and assigning corresponding confidence scores. Also, it predicts a group of offsets for each corner to remap the corners to the input image. A group of centripetal shifts are simultaneously predicted for each corner to represent a vector pointing from the corner to the center of the object, which are used to identify whether two corners are from the same object. The cross-star deformable convolution [35] is introduced to enrich the visual features of corner positions, whose offset field is learned from the guiding shift, which is a shift from the corner to the center of the object.

The bounding constrained center attention module is designed to predict an extra center point and a bounding constrained vector of the object, which can enforce the backbone network to simultaneously pay attention to the interior information of the object. This module is pruned after the framework training to ensure that it will not introduce an extra computational consumption during the detection.

## 2.2 Backbone network

Considering the limitation of the computational resource, the DLA [36], a lightweight deep learning network, is employed as the backbone network of the proposed CA-CentripetalNet, illustrated in Fig. 2. At the stage of Feature extraction, a series of convolution blocks and downsampling layers are utilized to extract image features with different resolutions and semantic levels. The feature extraction process can be formulated as

$$F_{i+1} = E_i(F_i) = E_i\left(E_{i-1}(......E_0(F_0))\right) \quad (1)$$

where $F_0$ is the input image and $F_i$ is the feature map of the $i$-th layer. $E_i$ denotes feature mapping from $F_i$ to $F_{i+1}$, which is composed of a series of convolutional layers, batch-norm layers and ReLU layers. The feature maps of the last three levels $\{F_4, F_5, F_6\}$ are selected as the output of the feature extraction stage, whose size are $\frac{H}{4} \times \frac{W}{4} \times 64$, $\frac{H}{8} \times \frac{W}{8} \times 128$ and $\frac{H}{16} \times \frac{W}{16} \times 256$, correspondingly.

At the feature aggregation stage, the lower-level feature maps are iteratively combined into the higher-level ones. Thus, a single group of feature maps can be achieved by this multi-input-single-output structure, which will be input into the prediction head. The feature aggregation process can be formulated as

$$F_{out} = \mathfrak{c}\{\mathfrak{c}[T(F_5), F_4], T[\mathfrak{c}(T(F_6), F_5)]\} \quad (2)$$

where $T()$ is a combination of a convolutional block and a transpose convolutional layer, upsampling the high-level feature maps. $\mathfrak{c}(\bullet, \bullet)$ is a concatenation block for feature maps.

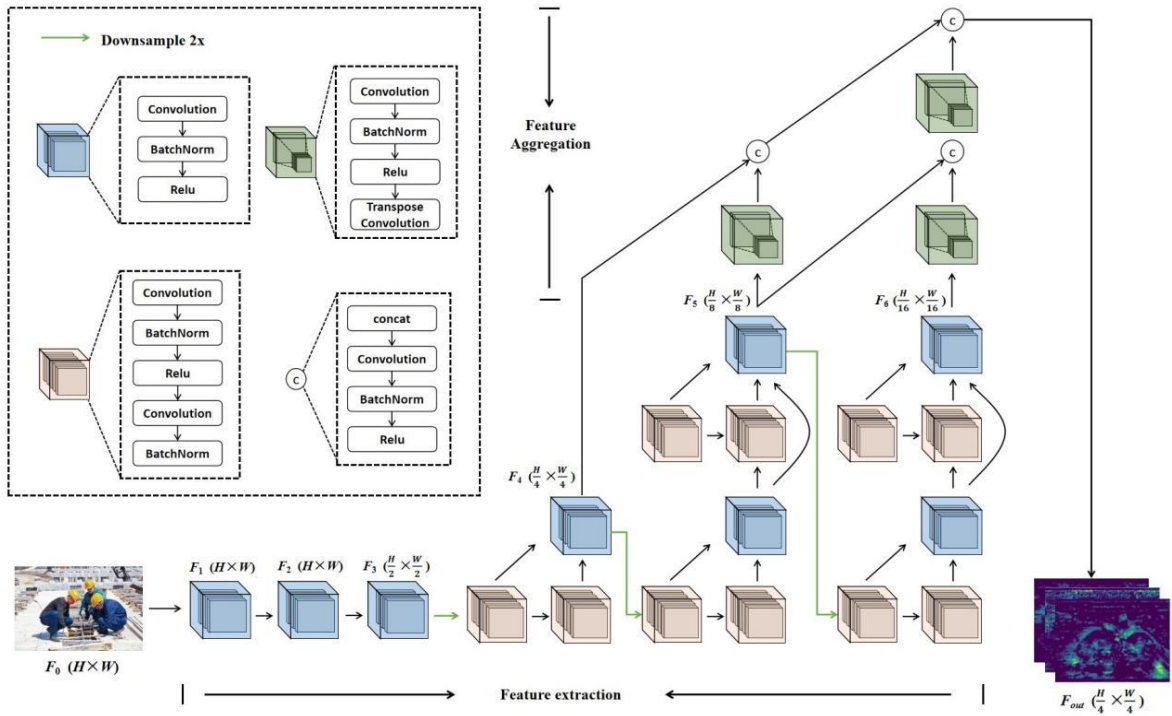

Fig. 2. The backbone network of the CA-CentripetalNet

## 2.3 Vertical-horizontal corner pooling

Corners are often outside the objects, which lack local appearance features. CornerNet [33] uses corner pooling to find the features on the boundary directions so as to determine corners. However, it makes corners sensitive to the edges.

Although the cascade corner pooling [34] can solve the problem of the sensitivity, its double corner pooling operation will result in some redundant semantic information since it make the corners "see" the interior of the objects in two directions at the same time. This will greatly influence the detection speed to limit its application in hardhat wearing detection.

For hardhat wearing detection, the horizontal (top and bottom) marginal information of the bounding box is more significant than the vertical (left and right) one. This is because a worn hardhat is usually on the top of the worker's head and the bottom marginal information can indicate whether there is a human neck. Comparatively, the vertical marginal information involves more background information, which is of less significance for detection. Thus, we design the vertical-horizontal corner pooling to focus on horizontal marginal features and internal features of the bounding box, which is beneficial for subsequent corner prediction. Fig. 3 shows the structure of the vertical-horizontal corner pooling.

Specifically, a Conv-BN-ReLU block is used to extract the features of the interior, horizontal margin, and corner of the object from the feature maps output by the backbone network. The maximal interior feature is focused on the horizontal margin by a vertical pooling, which is utilized to enhance the horizontal margin features by combining it with the feature maps output by the Conv-BN-ReLU block. The maximum of combined feature maps with the enhanced horizontal margin features are further focused on the corner by a horizontal pooling, which is utilized to enhance the conner features by the combination of the feature maps as illustrated in Fig. 3. Thus, the corners of the object aggregate the features of the interior, horizontal margin, and corner.

Assume $F_{pi}$ and $F_{po}$ as the input and output feature maps of the vertical and horizontal pooling operations, respectively, the pooling operations for the top-left corner prediction can be expressed as top pooling and left pooling, respectively, defined as

$$\text{Top Pooling: } F_{po_{ij}} = \begin{cases} \max\left(F_{pi_{ij}}, F_{po_{(i+1)j}}\right) & \text{if } i < H \\ F_{pi_{ij}} & \text{if } i = H \end{cases} \quad (3)$$

$$\text{Left Pooling: } F_{po_{ij}} = \begin{cases} \max\left(F_{pi_{ij}}, F_{po_{i(j+1)}}\right) & \text{if } j < W \\ F_{pi_{ij}} & \text{if } j = W \end{cases} \quad (4)$$

where $H$ and $W$ are the height and width of the feature maps.

Similarly, for bottom-right corner prediction, they can be expressed as bottom pooling and right pooling, respectively, defined as

$$\text{Bottom pooling: } F_{po_{ij}} = \begin{cases} \max\left(F_{pi_{ij}}, F_{po_{(i-1)j}}\right) & \text{if } i > 0 \\ F_{pi_{ij}} & \text{if } i = 0 \end{cases} \quad (5)$$

$$\text{Right pooling: } F_{po_{ij}} = \begin{cases} \max\left(F_{pi_{ij}}, F_{po_{i(j-1)}}\right) & \text{if } j > 0 \\ F_{pi_{ij}} & \text{if } j = 0 \end{cases} \quad (6)$$

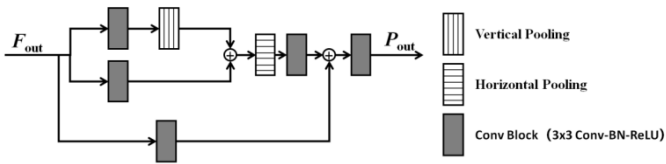

Fig. 3. The structure of vertical-horizontal corner pooling

### 2.4 bounding constrained center attention module

To enforce the backbone network to pay attention to the internal features of the object, a training module called bounding constrained center attention is designed as another network branch of the proposed framework, whose structure is illustrated in Fig. 4.

First, the center pooling [37] is designed to capture the most useful internal features from the feature maps output by the backbone network, which can find the maximal internal features in the horizontal/vertical directions by a Conv-BN-ReLU block, a left pooling and a right pooling and by a Conv-BN-ReLU block, a top pooling and a bottom pooling, respectively. The heatmaps of center points and a group of offsets for each point are predicted to represent the accurate positions and the categories of center points.

In addition, a bounding constraint prediction scheme is designed in the bounding constrained center attention module, which predicts a bounding constrained vector ($bc_h, bc_w$) for a center point of the object, defined as

$$bc_h = \log\left(\frac{br_x - tl_x}{2s}\right) \quad (7)$$

$$bc_w = \log\left(\frac{br_y - tl_y}{2s}\right) \quad (8)$$

where ($br_x$, $br_y$) and ($tl_x$, $tl_y$) are the coordinates of the bottom-right and top-left corners, respectively. $s$ denotes the downsample ratio of the backbone network.

## 3. Experimental results and discussions

### 3.1 Dataset and evaluation metrics

The GDUT-HWD dataset [19] are employed to validate our proposed framework. The samples in the dataset are divided into five categories (blue, red, white, yellow and non-hardhat-worn), covering large variations in construction sites, visual ranges, workers' posture, light background and occlusions. The dataset is divided into a training set, a validation set and a test set with the ratio of 5:2.5:2.5. To test the detection ability for objects with different scales, the samples are also divided into three scale categories, i.e. small objects (area ≤ $32^2$ pixels), medium objects ($32^2$ pixels ≤ area ≤ $96^2$ pixels) and large objects (area ≥ $96^2$ pixels).

Mean average precision (mAP) and the per-class average precision (AP) with an Interaction of Union (IoU) ≥ 0.5 are utilized to evaluate the detection performance of models.

### 3.2 Network implementation

All the models were implemented on a computer with an NVIDIA GTX1080ti GPU and an Intel Core i5-9400F @2.90GHz×6 CPU and in the MMDetection framework. We trained the proposed CA-CentripetalNet by the Adam optimizer with the batch of 6 and the learning rate of 0.0005 for 180 epochs and decayed the learning rate by a factor of 10 at the 150[th] epoch.

The objective function for the CA-CentripetalNet is defined as

$$L = L_{co}^{tl} + L_{co}^{br} + L_{bcca} \quad (9)$$

where the two $L_{co}$ denote the two losses for the top-left and bottom-right corners prediction modules, respectively. $L_{bcca}$ represents the loss for the bounding constrained center attention. They are defined as

$$L_{co} = L_{det}^{co} + L_{off}^{co} + L_{cs} + \alpha L_\delta \quad (10)$$

$$L_{bcca} = L_{det}^{ce} + L_{off}^{ce} + L_{ba} \quad (11)$$

where $L_{det}^{co}$ and $L_{det}^{ce}$ are Gaussian focal losses [33], which are used to train the framework to detect the corners and the center point of the object, respectively. $L_{off}^{co}$ and $L_{off}^{ce}$ denote the smooth L1 losses for the prediction of the offsets of the corners and the center point of the object, respectively. $L_{cs}$

and $L_\delta$ are smooth L1 losses for the prediction of the centripetal shift and the guiding shift of the cross-star deformable convolution, respectively. $L_{ba}$ denotes the smooth L1 loss for the prediction of the bounding constrained vector. $\alpha$ is a weight, which is empirically set to 0.05.

## 3.3 Influence of vertical-horizontal corner pooling

Here we conducted an experiment to discuss the influence of vertical-horizontal corner pooling (VHCP) on our proposed framework. Three pooling schemes were separately involved in the CA-CentripetalNet without bounding constrained center attention, which were corner pooling (CP) [33], cascade corner pooling (CCP) [34], and the proposed VHCP.

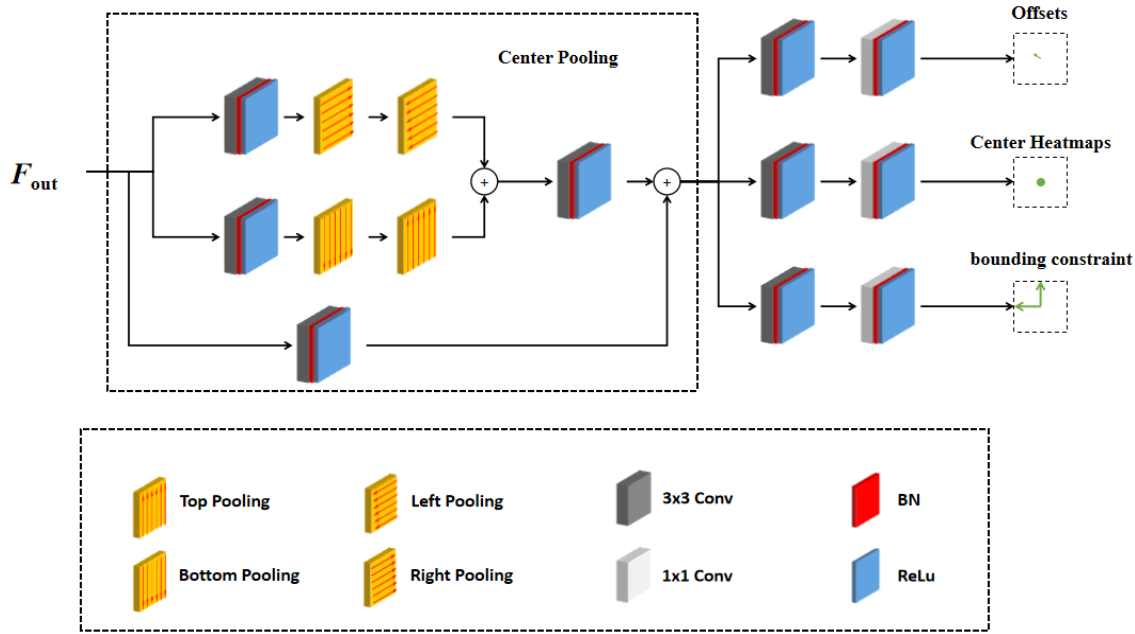

Fig. 4. The structure of the bounding constrained center attention.

As indicated in Table 1, the proposed framework with the CCP performs a little better detection than that with the CP. However, it has more network parameters and consumes more time for hardhat wearing detection compared with the two other models, which possibly limits its application in real construction sites. The framework with the VHCP can implement the hardhat wearing detection at the same speed as that with the CP and achieve similar performance as that with the CCP. This can be contributed to that the number of corner pooling operations for the VHCP is the same as that for the CP.

Table 1. performance of the proposed framework with different corner pooling.

| CP | CCP | VHCP | mAP(%) | size (MB) | FPS |
|----|-----|------|--------|-----------|------|
| √  |     |      | 87.52  | 33.5      | 26.6 |
|    | √   |      | 88.03  | 37.8      | 19.3 |
|    |     | √    | 87.98  | 34.7      | 26.6 |

Table 2. Performance of the proposed framework with/without the BCCA.

| Method | mAP(%) | size (MB) | FPS |
|--------|--------|-----------|------|
| without BCCA | 87.98 | 34.7 | 26.6 |
| with BCCA    | 88.63 | 34.7 | 26.6 |

## 3.4 Influence of bounding constrained center attention

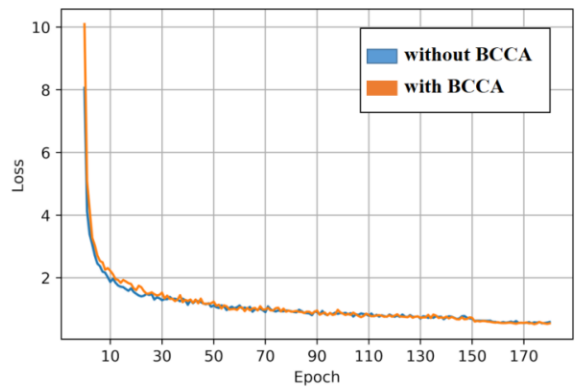

Fig. 5. Convergence of the proposed framework with/without the BCCA.

In this section, we conducted an experiment to demonstrate the contribution of the bounding constrained center attention (BCCA) to the proposed framework.

As shown in Fig. 5, the curves of the losses for the CA-CentripetalNet models with/without the BCCA almost overlap each other. The fact indicates that the BCCA does not bring much additional time cost during the training.

As indicated in Table 2, the proposed framework with the BCCA achieves better performance for hardhat wearing detection than that without BCCA. It is noted that the two models have the same size of network parameters and detection speed since the BCCA is a training module.

## 3.5 Comparisons with existing deep learning based methods

To validate the proposed hardhat wearing detection framework, we compared it with several deep learning based methods, such as MoblieNet-TD [21], Faster R-CNN [18], RetinaNet [37], SSD-RPA [19] and YOLOv5x [26].

As shown in Table 3, the MobileNet-TD [21] achieves the highest detection speed (54.6 FPS), mainly since its lightweight backbone network uses a lot of depthwise convolution and pointwise convolution. However, it has the poor ability of features extraction, resulting that it achieves the worst hardhat wearing detection accuracy of 75.56% mAP among all the methods, especially for small objects (59.91% mAP).

The Faster R-CNN [18] relies on the high-level and low-resolution features and is difficult to make full use of the low-level and high-resolution features, which not beneficial to characterize the small-scale objects. The RetinaNet [37] uses the ResNet as the backbone network and the FPN as the neck network, which is also weak in feature aggregation. Thus, the two methods perform bad performance for detecting the small-scale hardhats (61.11% mAP and 60.77% mAP, respectively). Conversely, they perform fairly good performance for large-scale hardhats also due to the above network characteristics, which are merely worse than our proposed method.

It is noted that the three above methods do not detect many workers without wearing hardhats, which is fatal for the safety management in construction sites. They only achieve less than 67% average precisions for non-hardhat wearing detection. This is possibly contributed to their weakness in feature aggregation.

The SSD-RPA [19] and the YOLOV5x [26] utilize different aggregation schemes to improve the detection ability. The former designs a reverse progressive attention to aggregate multi-level features for the objects. The latter integrates a series of effective network structures to improve the ability of feature extraction and aggregation. So, they greatly improve the detection ability for small-scale hardhats and non-worn-hardhats compared with the MobileNet_TD, the Faster R-CNN and the RetinaNet. However, they are anchor-based, whose limitation of the poor generalization results that, for example, some worn hardhats with relatively extreme scales are possibly mis-distinguished.

Among all the methods, our proposed framework achieves the best detection performance for the hardhats with different scales (80.75%, 94.85% and 95.04% mAPs for small-scale, medium-scale and large-scale hardhats, respectively), especially detect the most workers without wearing hardhats (84.48% average precision), at a reasonable speed of 26.6 FPS. This can be contributed to two reasons. One is that the proposed framework is an anchor-free detection model, which has the advantages of high efficiency, simple structure and rich generalization. The other is that the proposed schemes, such as the vertical-horizontal corner pooling and the bounding constrained center attention, effectively improves the ability of our model to extract and utilize image features. What's more, our framework has the smallest model size due to the simple model structure, which is beneficial for practical applications.

Although our method implements the task of hardhat wearing detection at the speed of 26.6 FPS, it is conducted on the computer with an NVIDIA GTX1080ti GPU, which is possibly expensive for practical applications in construction sites. In the future, we will further optimize the structure of the proposed framework to make it more lightweight and implement the optimized framework in CPU rather than in GPU.

Table 3. Comparisons of different deep learning methods for the GDUT-HWD dataset.

| Methods | Input size | Size (MB) | FPS | mAP (%) | small | medium | large | Average Precision(%) | | | | |
|---|---|---|---|---|---|---|---|---|---|---|---|---|
| | | | | | | | | none | blue | red | white | yellow |
| MobileNet-TD | 512×512 | 62.4 | 54.6 | 75.56 | 59.91 | 88.59 | 87.75 | 62.42 | 80.46 | 80.41 | 75.88 | 78.61 |
| Faster R-CNN | 512×512 | 158.7 | 25.2 | 76.52 | 61.11 | 89.64 | 92.92 | 66.61 | 82.13 | 80.12 | 77.10 | 76.64 |
| Reinanet | 512×512 | 139.5 | 29.0 | 77.76 | 60.77 | 89.95 | 91.62 | 66.91 | 83.45 | 81.32 | 76.33 | 80.83 |
| SSD-RPA | 512x512 | 118.1 | 30.3 | 83.96 | 76.52 | 90.37 | 90.08 | 82.44 | 83.26 | 85.35 | 84.62 | 84.13 |
| YOLOv5x | 416x416 | 173.1 | **45.5** | 86.23 | 78.21 | 93.65 | 90.37 | 79.53 | **90.77** | 87.68 | 86.47 | 86.75 |
| Our method | 512×512 | **34.7** | 26.6 | **88.63** | **80.75** | **94.85** | **95.04** | **84.48** | 89.57 | **90.49** | **89.45** | **89.16** |

## 3.6 Visualization

Some detection examples are visually shown in Fig. 6, which are achieved by our proposed method on the GDUT-HWD dataset. These examples cover a variety of construction scenes, including different visual ranges, workers' postures, light backgrounds and occlusions. These visualization results indicate that our proposed method can excellently distinguish whether the workers are wearing hardhats and identify the color of worn hardhats, even when the workers work in the construction sites with complicated video surveillance scenes, such as crowds, distant view and dim background. The facts demonstrate the potential of our proposed method for hardhat wearing detection in real construction sites.

## 4 Conclusions

Hardhat is an important personal protective equipment for safety management in construction sites, which is beneficial to ensure the safety of workers, and even to save their lives from fatal injuries. Automatic hardhat wearing detection based on computer vision can improve the efficiency of safety management in construction sites, which attracts more and more researchers. However, it is still challenging due to the complicated video surveillance scenes, such as crowds, distant view and dim background.

To deal with the inherent problems existed in the previous anchor-based deep learning methods for hardhat wearing detection, we design a novel anchor-free deep learning framework for single-step hardhat wearing detection, which is called CA-CentripetalNet. In the CA-CentripetalNet, a lightweight deep learning network named DLA is employed as the backbone network, which has an excellent ability of feature aggregation. Two novel schemes are involved in the framework to improve the detection performance with a reasonable computational burden and a reasonable memory consumption, which are vertical-horizontal corner pooling and bounding constrained center attention. It is noted that the bounding constrained center attention only works during the training to enforce the backbone network to pay attention to the interior of the object.

The ablation study indicates that the two novel schemes can improve the detection performance without the increase of the memory consumption. The proposed CA-CentripetalNet can achieve good detection performance of the total 88.63% mAP at a reasonable speed of 26.6 FPS for hardhat wearing detection on the GDUT-HWD dataset. It performs better detection performance than the existing deep learning methods, especially in case of small-scale hardhats and non-worn-hardhats. It is noted that it consumes the least memory for the model among all the deep learning methods. The facts demonstrate the potential of the CA-CentripetalNet for automatic hardhat wearing detection in real construction sites.

## Acknowledgments

This work was in part supported by the National Natural Science Foundation of China (Nos. 61901123 and 62171142), the Research Fund for Colleges and Universities in Huizhou (No. 2019HZKY003), and the Project of Jihua Laboratory (No.X190071UZ190).

## Declarations

### Competing interests

No potential conflict of interest was reported by the authors.

### Authors' contributions

Zhijian Liu and Nian Cai proposed ideas and methodologies, designed the experimental scheme and wrote the main manuscript text. Wensheng Ouyang and Chengbin Zhang assisted in the experiment and prepared figures. Nili Tian and Han Wang reviewed and modified the manuscript.

### Funding

No funding is received.

### Availability of data and materials

The GDUT-HWD dataset [19] can be accessed at https://github.com/wujixiu/helmet-detection/tree/master/hardhat-wearing-detection.

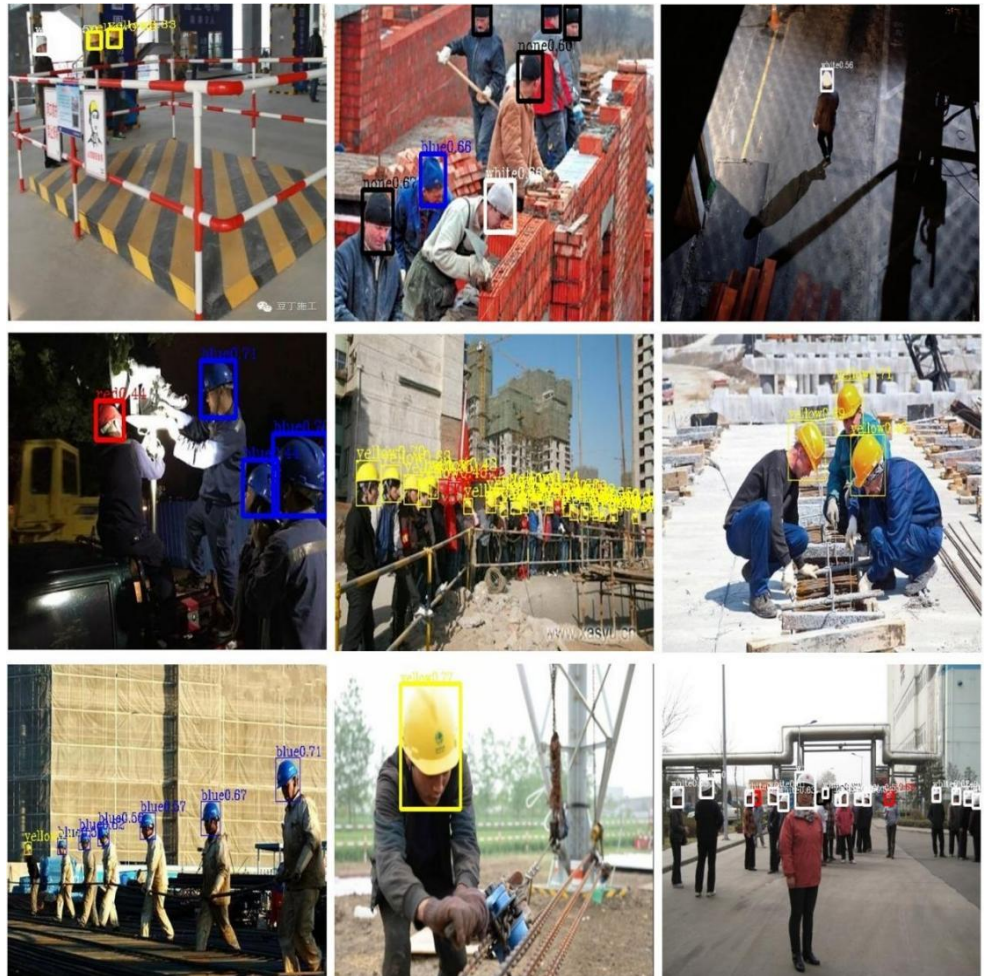

Fig. 6. Detection examples on GDUT-HWD with the proposed method.